\documentclass{article}

\usepackage{PRIMEarxiv}

\usepackage[utf8]{inputenc} 
\usepackage[T1]{fontenc}    
\usepackage{hyperref}       
\usepackage{url}            
\usepackage{booktabs}       
\usepackage{amsfonts}       
\usepackage{nicefrac}       
\usepackage{microtype}      
\usepackage{lipsum}
\usepackage{fancyhdr}       
\usepackage{graphicx}       
\usepackage{amsmath}
\usepackage{algorithm}
\usepackage{algorithmic}
\usepackage{amsthm}  
\usepackage{wrapfig}
\usepackage{multirow}
\theoremstyle{definition}
\newtheorem{defn}{Definition}[section]
\graphicspath{{media/}}     

\title{PRISM: Structured Optimization via Anisotropic Spectral Shaping}

\author{
  Yujie Yang \\
  \texttt{yangyj.m@outlook.com} \\
}

\begin{document}
\maketitle

\begin{abstract}
We propose PRISM, an optimizer that enhances first-order spectral descent methods like Muon with partial second-order information. It constructs an efficient, low-rank quasi-second-order preconditioner via innovation-augmented polar decomposition. This mechanism enables PRISM to perform anisotropic spectral shaping, which adaptively suppresses updates in high-variance subspaces while preserving update strength in signal-dominated directions. Crucially, this is achieved with minimal computational overhead and zero additional memory compared to first-order baselines. PRISM demonstrates a practical strategy for integrating curvature-adaptive properties into the spectral optimization paradigm.
\end{abstract}


\section{Introduction}

The remarkable success of large-scale models is critically dependent on the optimizer's ability to navigate high-dimensional, non-convex, and often ill-conditioned loss landscapes. Consequently, the focus of optimization research has shifted from the simple steepest descent of first-order methods towards more sophisticated strategies that approximate and exploit the underlying geometric structure of the parameter space.

A dominant family is adaptive optimization, such as AdaGrad \cite{duchi2011adaptive}, RMSprop \cite{tieleman2012rmsprop} and Adam \cite{kingma2014adam}. These optimizers construct a diagonal preconditioning matrix using historical gradient statistics, typically the running second moments. This yields a powerful mechanism for per-parameter learning rate scaling, which effectively normalizes the update magnitudes along each coordinate. However, the structure of this diagonal preconditioner imposes a fundamental limitation: it operates on parameters element-wise, rendering it intrinsically blind to the structural correlations and off-diagonal curvature information inherent in the weight matrix.

To address this, two primary research lines have emerged to incorporate the inherent structure of weight matrix.
The first one enhances the preconditioner of adaptive methods with richer structures. Methods like Shampoo \cite{gupta2018shampoo} and SOAP \cite{vyas2024soap} extend the adaptive philosophy from diagonal to more expressive Kronecker-factored structures. This approach captures row and column statistics separately, bringing the optimizer theoretically closer to second-order Newton methods \cite{morwani2024new}.
The other refines the update direction using principles of spectral descent, pioneered by methods like SSD \cite{carlson2015preconditioned} and made practical for large-scale models by Muon \cite{jordan6muon}. It reframes the problem and seeks the steepest descent direction under a spectral norm, which respects the nature of matrices as linear operators. 
By orthogonalizing the momentum matrix with Newton-Schulz iterations instead of SVD, Muon has successfully scaled structured optimization to large-scale model training \cite{liu2025muon}.

The success of structured methods highlights a shared principle: respecting the geometry of model parameters is critical for performance. 
Although computationally efficient with strong geometric intuition, the Muon optimizer also reveals the next frontier for improvement. 
As a spectral method, Muon applies an isotropic correction to the gradient by whitening the spectrum of the momentum matrix, thereby normalizing all its singular values, but it lacks the anisotropic adaptivity required by complex loss landscapes. 
This limitation stems from a fundamental statistical blindness within Muon. The preconditioner implicitly used by Muon, $P = (M^\top M)^{-1/2}$, is constructed solely from the first moment of the gradients.
An ideal preconditioner, such as the Fisher Information Matrix, is based on the second moment, $\mathbb{E}[G^\top G]$, which includes the gradient covariance. By omitting this covariance term, Muon may make overly aggressive updates in high-variance subspaces, such as loss canyons, potentially leading to instability.

In this work, we bridge this gap by introducing PRISM (PReconditioned Innovation-augmented spectral Shaping Momentum) optimizer that integrates partial second-order information into the spectral descent framework. Our core contribution is a mechanism we term innovation-augmented polar decomposition, which uses the innovation, the difference between the observed gradient and its momentum-based prediction ($D_t = G_t - M_t$), as an efficient, low-rank proxy for the gradient covariance. By orthogonalizing the innovation-augmented momentum matrix $\tilde{M} = \begin{bmatrix} M \\ \gamma D \end{bmatrix}$, PRISM acts as a implicit covariance-aware preconditioner:
$$
P_t = (M_t^\top M_t + \gamma^2 D_t^\top D_t)^{-1/2} \approx (\mathbb{E}[G]^\top \mathbb{E}[G] + \underbrace{\text{Cov}(G)}_{\text{low-rank approx}})^{-1/2}
$$
This design transforms the update mechanism from isotropic spectral whitening to anisotropic spectral shaping. PRISM automatically applies strong damping in subspaces with high gradient variance while maintaining aggressive updates in directions with a consistent signal. Crucially, this elevates the spectral descent paradigm to a quasi-second-order method that is aware of gradient uncertainty, with negligible computational overhead and no additional memory cost.

\section{Related Work}

The evolution of deep learning optimizers has two primary distinct methodologies: one focuses on utilizing gradient statistics for adaptive step-size adjustment, while the other aims to exploit the global structure of parameters to realize geometric optimization. 

\subsection{Adaptive Optimization}

The core premise of these method is to approximate the local curvature of the loss landscape with gradient statistics, thereby enabling independent scaling of update steps for each parameter. Algorithms such as AdaGrad \cite{duchi2011adaptive}, RMSProp \cite{tieleman2012rmsprop}, and Adam \cite{kingma2014adam} have established the de facto standard for training deep networks.
Adam, in particular, computes individual adaptive learning rates for each parameter, effectively implementing preconditioning by using a running average of the squared gradients as a computationally efficient approximation of the Hessian matrix's diagonal.
 
Although adaptive optimizers approximate local curvature and enable elsment-wise step size adjustment, these methods treat the parameter as a flatten vector with independent neurons, ignoring the structural correlations within weight matrices.

\subsection{Structured Optimization}

To overcome the geometric limitations of element-wise methods, structured optimization explicitly treats neural network weights as linear operators (matrices or tensors), applying structured theory to guide the update process.

SSD \cite{carlson2015preconditioned} pioneers the concept of spectral descent in deep learning, positing that for matrix parameters, the steepest descent direction under a spectral norm constraint is superior to that under the Euclidean ($L_2$) norm. This approach effectively discards the singular values of the gradient matrix, ensuring that all of its corresponding singular subspaces are updated with a uniform magnitude. 
Recognizing the need for further adaptivity, the concurrent RMSspectral and ADAspectral \cite{carlson2015preconditioned} methods straightforwardly combined the spectral update with element-wise scaling, effectively blending the two optimization paradigms.

Shampoo \cite{gupta2018shampoo} advances the structured approach by constructing a preconditioner that approximates the full-matrix AdaGrad using Kronecker products of rows and columns statistics.
Recent analysis reveals that without temporal accumulation, the Shampoo optimizer's update rule is equivalent to the steepest descent direction under the spectral norm \cite{bernstein2024old}. From this perspective, the standard Shampoo maintains the geometry of spectral optimization while leveraging statistics to achieve a superior second-order approximation.
Despite its theoretical completeness, additional memory and the computational cost of calculating inverse-fourth roots restricts its large-scale application.

Muon \cite{jordan6muon} inherits the geometric perspective of SSD, but revolutionizes the implementation. Instead of costly SVD, Muon employs Newton-Schulz iterations to approximate the polar decomposition, which is a highly efficient algorithm via recursive matrix multiplications:
$$
X_{k+1} = a X_k + b X_k(X_k^\top X_k) + c X_k(X_k^\top X_k)^2
$$
This method allows for the rapid orthogonalization of the momentum matrix on modern GPUs.

\section{Methodology}

\subsection{A Preconditioning Perspective for Muon Optimizer}

The Muon update is driven by the orthogonal polar factor $U$ of the momentum matrix $M$. This factor has two equivalent algebraic forms:
$$
\text{polar}(M) = U = (M M^\top)^{-1/2} M = M (M^\top M)^{-1/2}
$$
This duality implies that the Muon update can be interpreted interchangeably as either a left-preconditioned descent using row correlations ($M M^\top$) or a right-preconditioned descent using column correlations ($M^\top M$).

Adopting the right-preconditioned view, the Muon update direction is formed by the preconditioner $P=(M^\top M)^{-1/2}$ derived from the momentum $M$. Since the momentum $M$ is an exponential moving average that estimates the first moment of the gradient, $\mathbb{E}[G]$, the preconditioner is effectively built from the outer product of this mean:
$$
P^{\text{Muon}} = (\mathbb{E}[G]^\top \mathbb{E}[G])^{-1/2}
$$
This reveals a critical statistical deficiency. An ideal preconditioner, which approximates the curvature of the loss landscape
should be based on the second raw moment $\mathbb{E}[G^\top G]$. 
As the second moment decomposes into $\mathbb{E}[GG^\top] = \mathbb{E}[G]\mathbb{E}[G]^\top + \text{Cov}(G)$.
Muon's design, based solely on the first moment, inherently ignores the covariance information. As a result, the optimizer is blind to gradient variance, leading it to take aggressive and potentially unstable steps in uncertain direction.

\subsection{Innovation-Augmented Polar Decomposition}

To bridge this statistical gap, we propose innovation-augmented polar decomposition method. Our objective is to recover the missing covariance term $\text{Cov}(G)$ of the preconditioner via a structural augmentation of the orthogonalized variables.
\begin{defn}
  \label{def:inj}
\textbf{Innovation-Augmented Momentum.} At each step $t$, we define the instantaneous innovation term $D_t = G_t - M_t$, which serves as a rank-1 proxy for the gradient covariance. We construct an innovation-augmented momentum matrix $\tilde{M}_t$ by concatenating the original momentum $M_t$ and the scaled innovation $D_t$ along the primary dimension:
$$
\tilde{M}_t = \begin{bmatrix} M_t \\ \gamma D_t \end{bmatrix} \in \mathbb{R}^{2m \times n}
$$
where $\gamma$ is a damping coefficient controlling the sensitivity.
\end{defn}

We then calculate the polar decomposition of this augmented system, $\tilde{O}_t = \text{polar}(\tilde{M}_t)$. The resulting orthogonalized matrix can be expressed in block form as:
$$
\begin{aligned}
\tilde{O}_t &= \tilde{M}_t (\tilde{M}_t^\top \tilde{M}_t)^{-1/2} \\
&= \begin{bmatrix} M_t \\ \gamma D_t \end{bmatrix} (M_t^\top M_t + \gamma^2 D_t^\top D_t)^{-1/2} \\
\end{aligned}
$$

The update is obtained by slicing the top block of $\tilde{O}_t$. As shown in the equation above, this effectively applies a modified preconditioner to the momentum:

$$
O_t = \tilde{O}_t[:m, :] = M_t (M_t^\top M_t + \gamma^2 D_t^\top D_t)^{-1/2}
$$ 

This derivation demonstrates that orthogonalizing the augmented matrix is mathematically equivalent to optimizing with a implicit curvature-aware preconditioner $P_t^{\text{PRISM}}=(M_t^\top M_t + \gamma^2 D_t^\top D_t)^{-1/2} $. The term $\gamma^2 D_t^\top D_t$ acts as a low-rank correction, reintroducing the covariance information into the denominator.

We name the algorithm PRISM optimizer; see Algorithm~\ref{alg:prism} for details.

\begin{algorithm}[h]
\caption{PRISM Optimizer}
\label{alg:prism}
\begin{algorithmic}[1]
\REQUIRE Parameters $\theta$, Learning Rate $\eta$, Momentum Coefficient $\beta$, Damping Coefficient $\gamma$
\STATE Initialize $M_0$
\FOR{$t = 1, 2, \dots$}
    \STATE $G_t \leftarrow \nabla_{\theta} \mathcal{L}(\theta_{t-1})$
    \STATE $M_t \leftarrow \beta M_{t-1} + (1 - \beta)G_t$
    \STATE $D_t \leftarrow G_t - M_t$
    \STATE $\tilde{M}_t \leftarrow \text{Concat}([M_t; \gamma D_t])$ 
    \STATE $\tilde{O}_t \leftarrow \text{polar}(\tilde{M}_t)$ \COMMENT{e.g., via NS iteration}
    \STATE $O_t \leftarrow \tilde{O}_t[:m, :]$
    \STATE $\theta_{t} \leftarrow \theta_{t-1} - \eta \cdot O_t$
\ENDFOR
\end{algorithmic}
\end{algorithm}

\subsection{Anisotropic Spectral Shaping}

In this section, we analyze the PRISM update mechanism from a spectral perspective. Using an eigenspace analysis, we demonstrate how PRISM surpasses the isotropic spectral whitening limitations of its predecessors by achieving anisotropic spectral shaping that dynamically modulates the update magnitude in each principal direction based on a local SNR.

\subsubsection{Eigenvalue Decomposition: Signal vs. Noise Energy}

Since the innovation-augmented Gram matrix $G_{\text{aug}} = \tilde{M}_t^\top \tilde{M}_t = M_t^\top M_t + \gamma^2 D_t^\top D_t$ is a real, symmetric, and positive-definite matrix, the spectral theorem guarantees its eigendecomposition, $G_{\text{aug}} = V \Lambda V^\top$. Here, $V = [v_1, \dots, v_n]$ is an orthonormal basis of eigenvectors defining the principal axes of the augmented momentum, and $\Lambda = \text{diag}(\lambda_1, \dots, \lambda_n)$ contains the corresponding eigenvalues.

Each eigenvalue $\lambda_k$ represents the total energy of the augmented momentum along its corresponding principal direction $v_k$. Applying the Rayleigh quotient yields a critical decomposition:
$$
\lambda_k = v_k^\top G_{\text{aug}} v_k = v_k^\top (M_t^\top M_t) v_k + \gamma^2 v_k^\top (D_t^\top D_t) v_k
$$
This simplifies to a sum of two distinct energy components:

$$
\lambda_k = \|M_t v_k\|^2 + \gamma^2 \|D_t v_k\|^2
$$
\paragraph{Signal Energy ($\|M_t v_k\|^2$)} The squared magnitude of the momentum (mean gradient) projected onto the eigenvector $v_k$. This term captures the consistent trend in that direction.
\paragraph{Noise Energy ($\|D_t v_k\|^2$)} The squared magnitude of the innovation (gradient variance) projected onto $v_k$. This term captures the stochasticity or uncertainty in that direction.

\subsubsection{Anisotropic Spectral Gain}
\label{sec:Spectral Gain}

The PRISM update, $O_t = M_t P_t = M_t V \Lambda^{-1/2} V^\top$, can be understood as an adaptive spectral filter. To show this, we analyze its effect on the momentum component within each eigenspace spanned by $v_k$. The update magnitude in this direction is modulated by a spectral gain, $\rho_k$:
$$
\|O_t v_k\| = \| M_t V \Lambda^{-1/2} V^\top v_k \| = \lambda_k^{-1/2} \|M_t v_k\|
$$
Substituting the energy decomposition into $\lambda_k$, we get the explicit form of this gain:
$$
\rho_k = \frac{\| M_t v_k \|}{\sqrt{\| M_t v_k \|^2 + \gamma^2 \| D_t v_k \|^2}}
$$
This gain coefficient, $\rho_k$, acts as an adaptive gain. It can be expressed directly in terms of the Signal-to-Noise Ratio (SNR) for the $k$-th principal direction, where $\text{SNR}_k = \frac{\|M_t v_k\|}{\gamma \|D_t v_k\|}$:
$$
\rho_k = \frac{1}{\sqrt{1 + 1/\text{SNR}_k^2}}
$$

\subsubsection{Analysis of the Spectral Gain}

This SNR-driven behavior creates two distinct operational regimes:

\paragraph{Case I: High-SNR Subspace ($\text{SNR}_k \gg 1$)}
In directions where the gradient signal is consistent, the noise energy is negligible ($\|D_t v_k\| \approx 0$). The SNR is high, and the spectral gain approaches unity:
$$
\rho_k \to 1
$$
Here, PRISM mimics spectral whitening optimizers like Muon, normalizing the update magnitude ($\|O_t v_k\| \approx \|M_t v_k\| / \|M_t v_k\| = 1$). This enables confident, aggressive steps along directions with a clear, stable signal.

\paragraph{Case II: Low-SNR Subspace ($\text{SNR}_k \ll 1$)}
In directions characterized by high variance, the noise energy $\gamma^2 \|D_t v_k\|^2$ dominates. The SNR is low, introducing a strong damping effect on the gain:
$$
\rho_k \approx \text{SNR}_k = \frac{\| M_t v_k \|}{\gamma \| D_t v_k \|} \ll 1
$$
Here, the update is heavily attenuated. This mechanism acts as a soft-threshold, effectively suppressing updates in directions where high gradient variance indicates uncertainty.

Through this mechanism, PRISM achieves anisotropic spectral shaping. It functions as an adaptive low-pass filter in the spectral domain, leveraging the innovation term to distinguish between the underlying signal and stochastic noise. Unlike methods that whiten the spectrum isotropically, PRISM selectively and proportionally damps updates in directions with less confidence.

\subsection{Memory and Computational Analysis}

\subsubsection{Memory Efficiency}

A key advantage of PRISM is its stateless design regarding second-order information. Unlike Shampoo, which requires storing historical Gram matrices ($O(n^2+m^2)$), PRISM computes the innovation term $D_t$ and the augmented matrix $\tilde{M}_t$ on-the-fly. As a result, PRISM maintains the same memory footprint as Muon or other first-order optimizer, requiring no persistent storage beyond the first-moment accumulator.

\subsubsection{Computational Complexity}

Analogous to Shampoo, PRISM can be configured for either left- or right-sided preconditioning by changing the direction of augmentation. Let the gradient be $G \in \mathbb{R}^{m \times n}$ with $m \ge n$.

\textbf{Right-Sided Preconditioning:} This configuration, which targets correlations between the $n$ columns, is achieved by concatenating the innovation along the row dimension, yielding an augmented matrix $\tilde{M} \in \mathbb{R}^{2m \times n}$. The computational cost is determined by the size of the smaller Gram matrix, $\tilde{M}^\top \tilde{M}$, which is $n \times n$. The cost of the single Newton-Schulz step is therefore:
$$
\mathcal{C}_{\text{Right}} \approx \mathcal{O}(4mn^2 + n^3)
$$

\textbf{Left-Sided Preconditioning:} To target correlations between the $m$ rows, we concatenate along the column dimension, forming $\tilde{M}_{\text{L}} = [M_t, \gamma D_t] \in \mathbb{R}^{m \times 2n}$. The corresponding left-sided preconditioner is $P_{\text{Left}} = (\tilde{M}_{\text{L}} \tilde{M}_{\text{L}}^\top)^{-1/2}$. The NS iteration can be efficiently applied to the smaller of the two possible Gram matrices: $\tilde{M}_{\text{L}}^\top \tilde{M}_{\text{L}}$ ($2n \times 2n$) or $\tilde{M}_{\text{L}} \tilde{M}_{\text{L}}^\top$ ($m \times m$). The complexity is thus dominated by $\min(m, 2n)$.

Compared to the baseline Muon optimizer, PRISM incurs only a minimal constant-factor increase in computational load primarily an approximate $2\times$ increase in the matrix multiplication operations which is highly amenable to efficient execution on modern hardware. Given that the Newton-Schulz iteration typically accounts for a minute fraction of the total training pipeline, the impact of this additional computation on the overall training throughput is negligible. 

Crucially, by strictly reusing the numerically stable and efficient Newton-Schulz iteration for the polar decomposition, our method ensures practicality and robustness.

\section{Experiments}
\subsection{Experimental Setup}
\subsubsection{Model Architecture and Dataset}

We evaluate PRISM on the task of causal language model pre-training. This standard benchmark, with its well-established model architectures and data pipelines, allows us to isolate and rigorously assess the optimizer's performance.

Our model is a decoder-only Transformer based on the Qwen2 \cite{team2024qwen2} architecture, comprising approximately 22M non-embedding parameters. We pre-train on the 10B token subset of the FineWeb-Edu dataset \cite{penedo2024fineweb}, consuming a total of 2.6B tokens for our experiments. The corpus is tokenized using the Qwen2 tokenizer with a fixed context length of 2048.

\subsubsection{Implementation Details and Hyperparameters}

Following the mixed-optimization paradigm of Muon, we partition model parameters into two groups:

\paragraph{Structured Parameters.} All high-dimensional weight matrices (e.g., linear projection in attention and MLP layers) are updated using either PRISM or Muon. For these parameters, we use a momentum coefficient of $\beta=0.95$ with Nesterov acceleration and adopt the learning rate scaling strategy from Moonlight.

\paragraph{Unstructured Parameters.} Embedding layer, LayerNorm parameters, and biases are updated using AdamW with $\beta_1=0.9$ and $\beta_2=0.95$.

All experiments use a global batch size of 128 and a cosine annealing learning rate schedule with a 2000-step linear warmup. The maximum learning rate for PRISM and Muon is set to 0.02, decaying to a final value of 0.002. Both parameter groups share the same schedule. We apply a weight decay of 0.01 and a gradient clipping threshold of 10.0. This relatively high clipping threshold is intentionally chosen to expose potential instabilities.

\subsubsection{Baselines}

We compare PRISM against two strong baselines: AdamW and Muon.

\paragraph{AdamW.} We use a tuned maximum learning rate of 0.005, a standard choice for this optimizer on similar tasks.

\paragraph{Muon.} Muon serves as a critical baseline as it is algorithmically equivalent to PRISM with the innovation coefficient set to zero ($\gamma=0$). This comparison directly isolates the contribution of PRISM's innovation-augmented spectral shaping.

\subsection{Results}

\begin{figure}[ht]
  \begin{center}
    \centerline{\includegraphics[width=0.75\columnwidth]{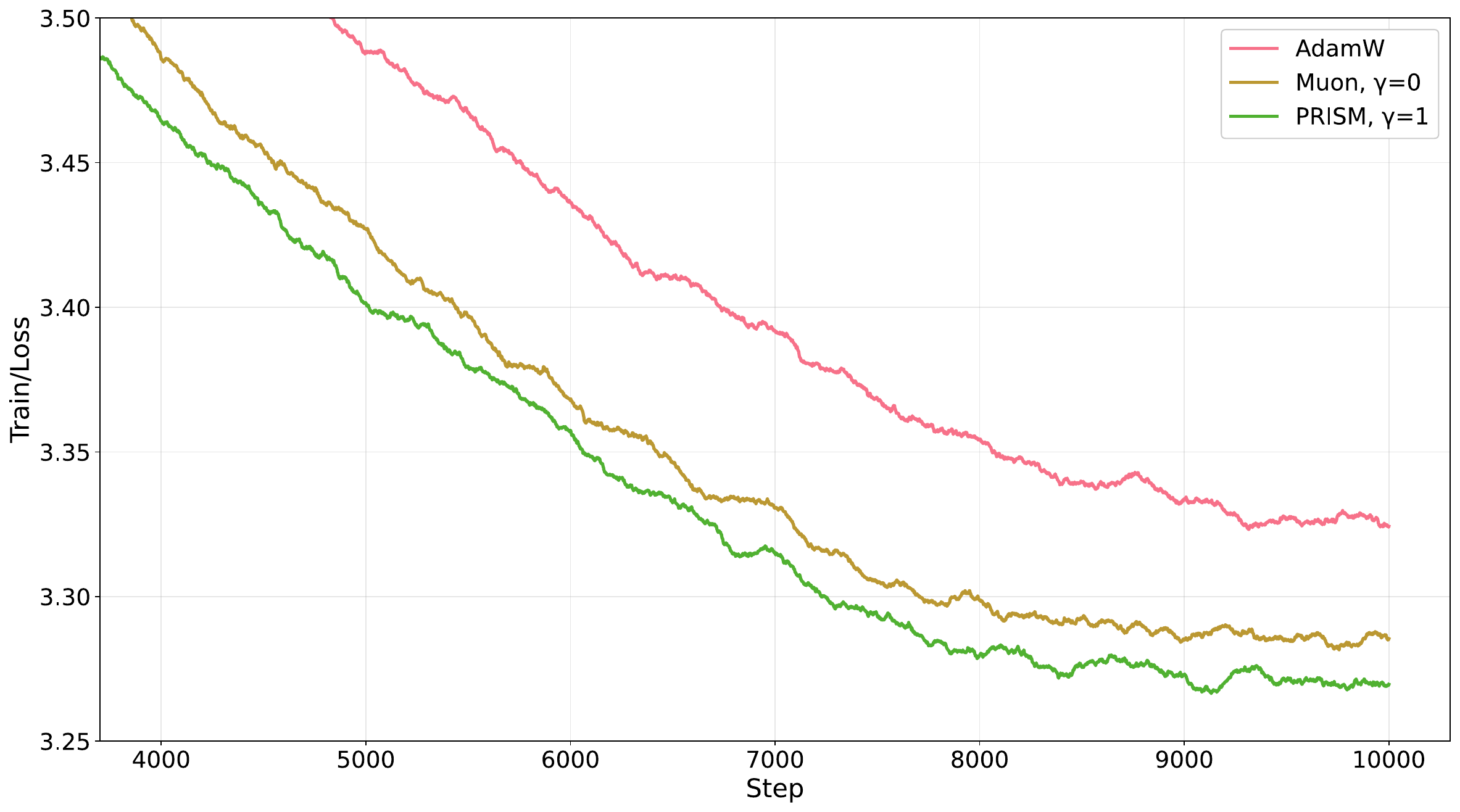}}
    \caption{
      Comparison of training loss curves for AdamW, Muon, and PRISM. 
    }
    \label{main_res}
  \end{center}
\end{figure}

Figure \ref{main_res} presents the training loss curves for AdamW, Muon, and PRISM ($\gamma=1$). The results confirm that structured optimizers (Muon and PRISM) substantially outperform AdamW, validating the advantage of leveraging the parameter matrix geometry. Building on this, PRISM consistently accelerates convergence over Muon. At 10,000 steps, PRISM achieves a final loss of 3.269, surpassing Muon’s 3.285. Like Muon, PRISM maintains excellent stability throughout training, exhibiting no loss spikes. This confirms that the innovation-augmented spectral shaping mechanism successfully improves convergence on this non-convex landscape.

\begin{table}[h]
\centering
\caption{Sensitivity analysis of the damping coefficient $\gamma$ on FineWeb-Edu pretraining tasks. Relative improvement is calculated against the Muon baseline.}
\label{tab:sensitivity}
\begin{tabular}{lccc}
\toprule
\textbf{Method} & \textbf{$\gamma$} & \textbf{Final Loss} & \textbf{Relative Improv.} \\
\midrule
AdamW & - & 3.324 & - \\
Muon (Baseline) & 0 & 3.285 & - \\
\midrule
PRISM & 0.1 & 3.290 & +0.005 \\
PRISM & 0.5 & 3.278 & -0.007 \\
PRISM & 1.0 & 3.269 & -0.016 \\
PRISM & 2.0 & 3.266 & -0.019 \\
PRISM & 5.0 & 3.266 & -0.019 \\
PRISM & 10.0 & 3.270 & -0.015 \\
\bottomrule
\end{tabular}
\end{table}

Table \ref{tab:sensitivity} summarizes the final training loss across a range of damping coefficients ($\gamma$). The data reveals that, with the exception of $\gamma=0.1$ where the damping effect is negligible, PRISM achieves performance superior to the baseline across a wide range of $\gamma \in [0.5, 10.0]$. This wide effective range demonstrates that the mechanism is robust and does not require precise hyperparameter tuning to deliver significant gains.

\subsection{Analysis}
\subsubsection{The Necessity of Anisotropic Shaping}

A core claim of our work is that PRISM's advantage stems from its anisotropic spectral shaping. To verify this, we ablate our design by comparing it against a simpler alternative: Muon augmented with standard Tikhonov damping.

The Tikhonov damping introduces a uniform damping term, $\lambda$, to the momentum Gram matrix, leading to the update rule:
$$ \text{Update} \propto M (M^\top M + \lambda I)^{-1/2} $$
The resulting spectral gain is $\rho_k = \|M v_k\| / \sqrt{\|M v_k\|^2 + \lambda}$. Critically, the damping factor $\lambda$ is a constant scalar, applied uniformly across all spectral directions. This acts as a crude low-pass filter, suppressing directions with low signal energy but remaining entirely blind to the underlying SNR of each direction.

\begin{figure}[ht]
\centering
\begin{minipage}[t]{0.48\linewidth}
  \centering
  \includegraphics[width=\linewidth]{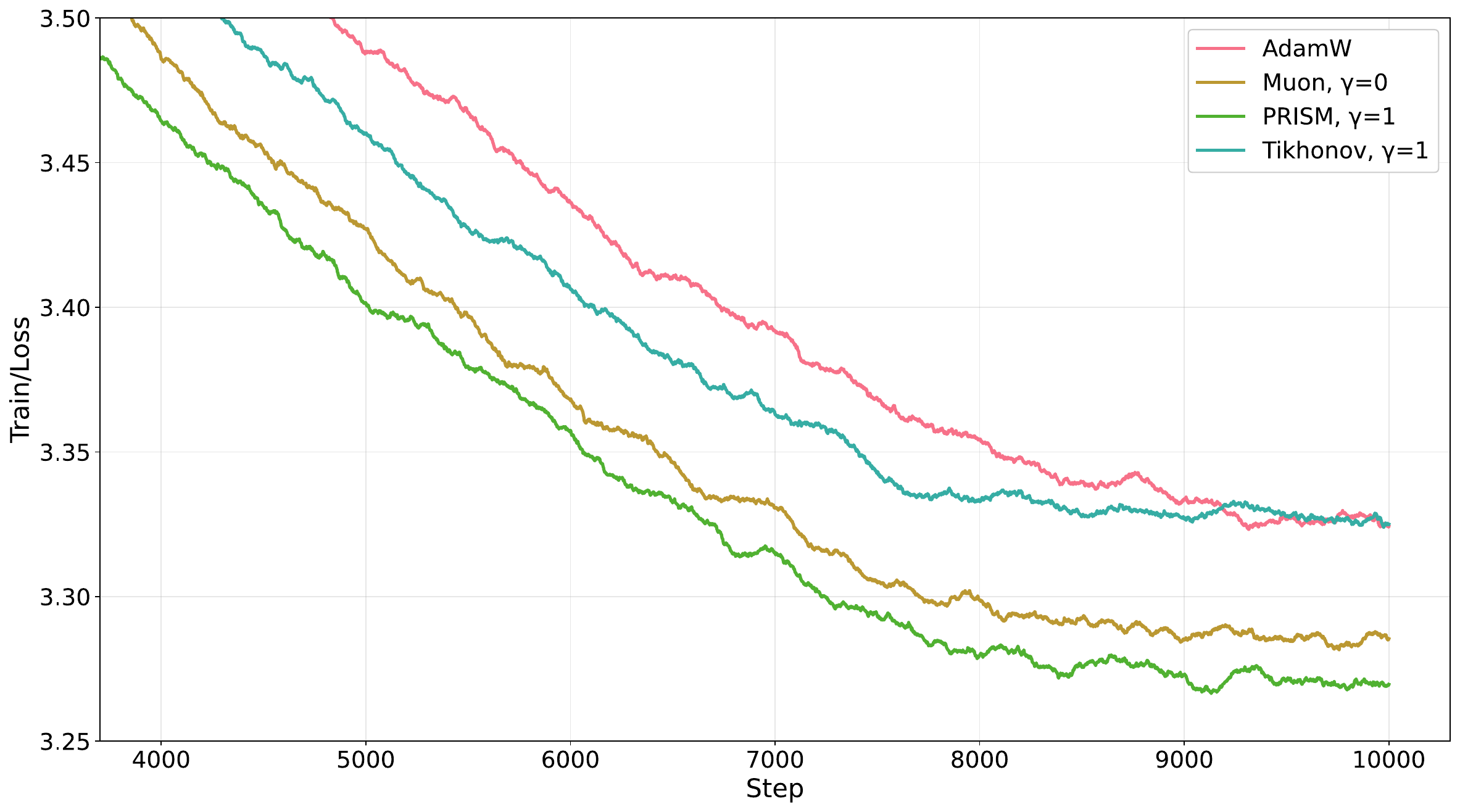}
  \caption{Training loss comparison between PRISM
           and Muon with Tikhonov damping.}
  \label{fig:tik_damping}
\end{minipage}\hfill
\begin{minipage}[t]{0.48\linewidth}
  \centering
  \includegraphics[width=\linewidth]{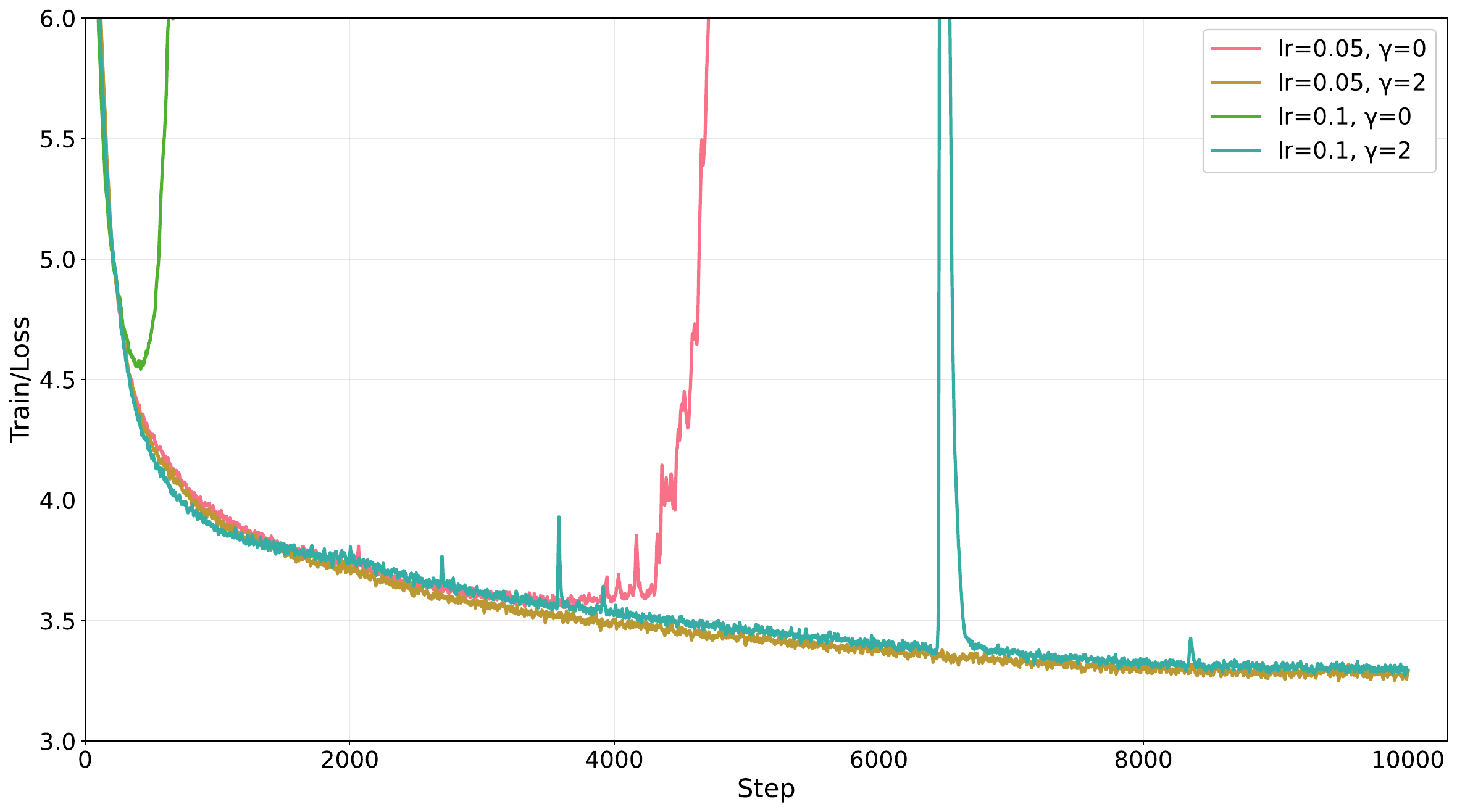}
  \caption{Comparison of training loss trajectories for Muon
           and PRISM under aggressive learning rates.}
  \label{fig:lr_aggressive}
\end{minipage}
\end{figure}

As shown in Figure \ref{fig:tik_damping}, Tikhonov-damped Muon achieves a final loss of 3.325, performing even worse than the undamped Muon baseline. This result is crucial: it demonstrates that simply regularizing or damping the update is insufficient. PRISM's superiority is not a consequence of mere regularization, but of its ability to selectively attenuate updates in noise-dominated subspaces while preserving full update strength in signal-dominated ones.

\subsubsection{Stability under Extreme Learning Rates}
We conducted stress tests with elevated learning rates to probe the optimizer's stability limits. As showed in Figure \ref{fig:lr_aggressive} and Table \ref{tab:high_lr_stability}, the Muon baseline diverges rapidly at learning rates of 0.05 and 0.1. In stark contrast, PRISM remains stable. Notably, at $lr=0.1$, PRISM's loss curve exhibits transient spikes but quickly self-recovers to convergence.

\begin{table}[h]
\centering
\caption{Stability comparison under aggressive learning rates. PRISM demonstrates a significantly wider safe operating region compared to Muon.}
\label{tab:high_lr_stability}
\begin{tabular}{lccc}
\toprule
\textbf{$lr$} & 0.02 & 0.05 & 0.1 \\
\midrule
$\gamma=0$ & 3.285 & Diverged & Diverged \\
$\gamma=2$ & \textbf{3.266} & \textbf{3.276} & \textbf{3.298} \\
\bottomrule
\end{tabular}
\end{table}

This self-recover property validates PRISM's dynamic regulation mechanism. When updates lead to regions of high loss curvature, the instantaneous innovation $D_t$ grows, triggering stronger, targeted damping that automatically stabilizes the training process. This demonstrates that PRISM significantly expands the optimizer's Safe Operating Region, enabling more aggressive and potentially faster training strategies.

\subsubsection{Internal Mechanisms of PRISM}

\begin{figure}[ht]
\centering
\begin{minipage}[t]{0.48\linewidth}
  \centering
  \includegraphics[width=\linewidth]{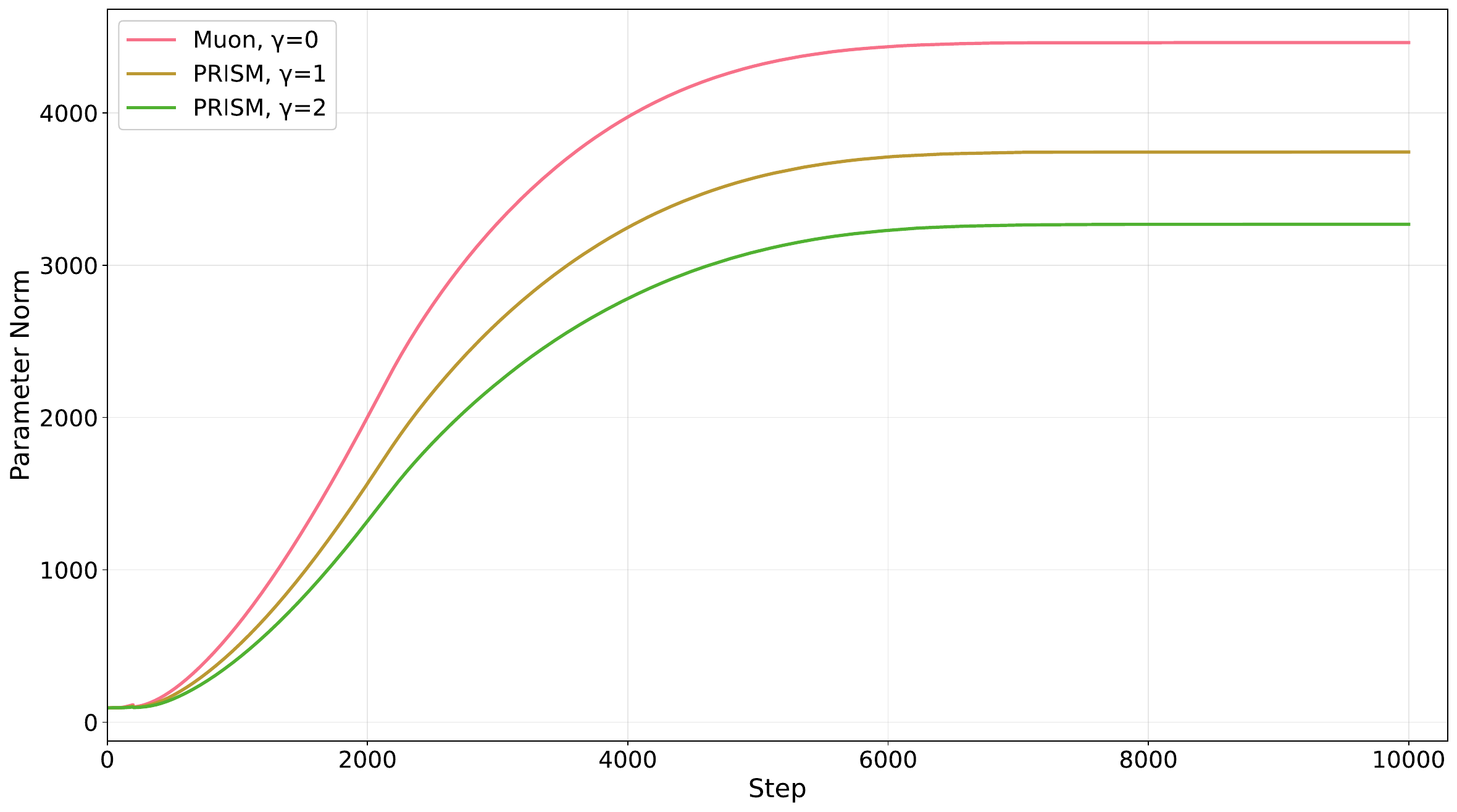}
  \caption{Frobenius norm of the parameters $\|O_t\|_F$ over training steps. }
  \label{fig:traning_norm}
\end{minipage}\hfill
\begin{minipage}[t]{0.48\linewidth}
  \centering
  \includegraphics[width=\linewidth]{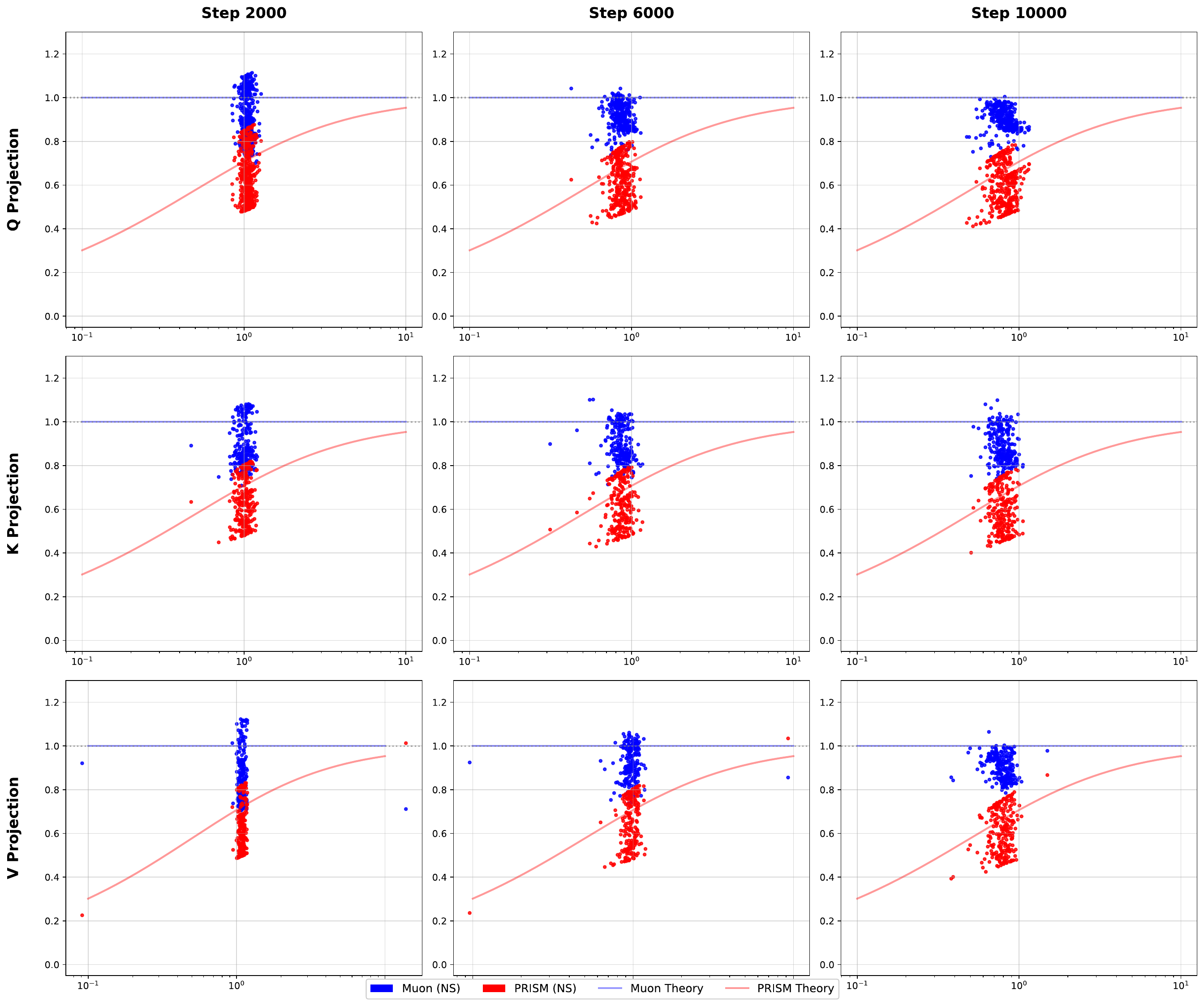}
  \caption{The relationship between the empirical SNR and the spectral gain ($\rho_k$) for subspaces.}
  \label{fig:spectral gain}
\end{minipage}
\end{figure}

\paragraph{Implicit Regularization.}
Figure \ref{fig:traning_norm} plots the Frobenius norm $\|O_t\|_F$ of the parameter matrix along training. The growth rate of the norm is inversely proportional to $\gamma$. By selectively suppressing updates in noisy directions, PRISM imposes an implicit regularization and reduce the update magnitude.

\paragraph{SNR-Aware Spectral Gain.}
Figure \ref{fig:spectral gain} provides a direct visual confirmation of our spectral analysis from section \ref{sec:Spectral Gain}. The spectral gain of Muon is around 1.0, irrespective of the SNR. The spectral gain of PRISM, however, follows the theoretical SNR-driven curve, $\rho_k = (1 + \gamma^2/\text{SNR}_k^2)^{-1/2}$. This empirically proves that the innovation-augmented update by NS iteration indeed shows theoretical behavior, performing adaptive, SNR-aware filtering.

Furthermore, the visualization reveals a temporal diffusion of the SNR distribution. As training progresses, the SNR values of different subspaces diverge. PRISM actively adapts to this evolution, whereas Muon remains blind, continuing to apply its rigid isotropic whitening.

\section{Conclusion}
We propose PRISM, an optimizer that addresses the statistical blindness of existing first-order spectral methods like Muon through a novel innovation-augmented spectral shaping mechanism. PRISM optimizer is able to adaptively damp updates in high-variance subspaces while preserve aggressive steps in stable ones. Empirically, PRISM demonstrates comprehensive advantages in large-scale language model pretraining. 
As a stateless and computationally efficient method, PRISM provides a mechanism for incorporating second-order information into spectral methods, bridging the two optimization perspectives.

\section{Future Work}
The PRISM framework establishes a novel paradigm for integrating quasi-second-order information into spectral optimization via innovation augmentation. While our current implementation prioritizes computational efficiency, the underlying principles suggest several promising avenues for future research.

A natural extension lies in generalizing beyond single-sided spectral shaping. While our current method defaults to a single-sided approach for efficiency—contrasting with the dual-sided preconditioners in Kronecker-factored methods—the optimal configuration remains an open question. Specifically, whether a dual-sided approach is superior \cite{xie2025structured}, and which dimension is optimal to target in the single-sided regime, are yet to be determined. This latter inquiry is motivated by our preliminary findings, where certain computationally intensive single-sided variants yielded observable performance improvements.

Another key frontier lies in enriching the covariance estimation. PRISM's instantaneous, rank-1 innovation is a deliberate design choice for maximal efficiency. A natural evolution is to develop methods that maintain a streaming, low-rank approximation of the gradient covariance via randomized matrix sketching techniques. Such a mechanism could provide a more robust signal for spectral shaping, further enhancing stability and performance in highly stochastic training regimes.

\bibliographystyle{unsrt}  
\bibliography{references}

\end{document}